\documentclass[12pt]{article}

\usepackage[utf8]{inputenc}
\usepackage[T1]{fontenc}
\usepackage{lmodern}
\usepackage[margin=2.5cm]{geometry}
\usepackage{graphicx}
\usepackage{booktabs}
\usepackage{array}
\usepackage{amsmath}
\usepackage{float}
\usepackage{caption}
\usepackage{subcaption}
\usepackage[numbers,sort&compress]{natbib}
\usepackage{url}
\usepackage{enumitem}
\usepackage{xcolor}
\usepackage{hyperref}
\usepackage{lineno}
\usepackage{times}
\usepackage{multirow}

\usepackage{adjustbox}

\usepackage{pifont}
\newcommand{\cmark}{\ding{51}} 
\newcommand{\xmark}{\ding{55}} 

\graphicspath{{figures/}}
\newcolumntype{P}[1]{>{\raggedright\arraybackslash}p{#1}}

\hypersetup{bookmarksopen=true,bookmarksopenlevel=2,
    colorlinks=true,
    linkcolor=blue!60!black,
    citecolor=blue!60!black,
    urlcolor=blue!60!black
}

\begin{document}
\sloppy

\title{A paired synthetic construction-site image dataset for robust computer vision under adverse conditions}

\author{
\parbox{0.88\textwidth}{
\centering
Viet Huy Duong\textsuperscript{1},
Ruoxin Xiong\textsuperscript{2,*},
Md Abdullah Al Forhad\textsuperscript{3},
Weishi Shi\textsuperscript{4}
}
}

\date{}
\maketitle

\noindent
\textsuperscript{1} Department of Computer Science, College of Arts and Sciences, Kent State University, Kent, Ohio, United States. Email: \texttt{vduong1@kent.edu}\\
\textsuperscript{2} Assistant Professor, Construction Management Program, College of Architecture and Environmental Design, Kent State University, Kent, Ohio, United States. Email: \texttt{rxiong3@kent.edu} (Corresponding author) \\
\textsuperscript{3} Department of Computer Science and Engineering, University of North Texas, Denton, Texas, United States. Email: \texttt{MdAbdullahAlForhad@my.unt.edu}\\
\textsuperscript{4} Assistant Professor, Department of Computer Science and Engineering, University of North Texas, Denton, Texas, United States. Email: \texttt{Weishi.Shi@unt.edu}\\

\begin{abstract} 
Computer-vision systems used for construction monitoring can degrade under adverse environmental and visual conditions, yet such conditions remain underrepresented in existing construction image datasets. We present \textit{ConSynth-X}, a paired synthetic construction-site image dataset containing 34,199 images derived from 3,109 real-world source scenes. The dataset comprises 11 condition-specific subsets spanning precipitation, fog, nighttime illumination, adverse weather at night, and small-object or long-distance views. Each synthetic image is linked to its corresponding source scene, enabling controlled comparison across environmental and visual conditions. ConSynth-X includes source-derived annotations, generation metadata, provenance information, and image-quality indicators, supporting object detection, image captioning, visual grounding, and visual question answering. Technical validation evaluates source--synthetic fidelity and alignment with real adverse-condition imagery using embedding-based similarity and distributional analyses. The dataset provides a structured resource for evaluating and improving the robustness of construction vision and vision--language models under challenging field conditions.
\end{abstract}

\section{Background \& Summary}
Construction field operations require timely and reliable information on workers, equipment, materials, site conditions, and work progress to support safety management, productivity assessment, progress monitoring, quality control, and project documentation~\cite{rao2022real,xiong2026openconstruction}. These information requirements are particularly important because construction activities are performed in dynamic environments characterized by substantial variation in weather, illumination, visibility, and viewing distance. In the United States, construction accounted for 20.8\% of occupational fatalities in 2023, with falls, slips, and trips accounting for 38.5\% of construction fatalities~\cite{BLS2025_FatalFallsConstruction2023}. Weather-related conditions have also been associated with construction injuries, fatalities, and productivity losses~\cite{sesesie2024exploring,larsson2023effects}, while nighttime operations introduce additional challenges associated with reduced and nonuniform illumination~\cite{nnaji2020effects}. Therefore, environmental and visual conditions represent important factors that affect construction operations and the reliability of technologies used for field monitoring.

Visual sensing has become an important means of acquiring construction-site information. Images and videos collected using fixed cameras, mobile devices, unmanned aerial systems, robotic platforms, and other visual sensors provide spatial and temporal observations of site activities and conditions~\cite{xiong2026openconstruction}. Computer-vision and vision-language methods can process these data to detect and localize workers, equipment, vehicles, materials, and personal protective equipment; characterize site activities and spatial relationships; generate image descriptions; and answer questions about construction scenes~\cite{xiao2022captioning,chen2025largepretrainedvisionlanguage}. The performance of these methods, however, is sensitive to the visual characteristics of the input data. Rain, snow, fog, low illumination, glare, shadows, nonuniform lighting, long viewing distances, and small object sizes can alter image contrast, visibility, appearance, and the amount of discriminative information available for visual recognition~\cite{ding2024robust,park2023small}. Consequently, performance measured primarily under clear and well-illuminated conditions may not adequately characterize model reliability across the broader range of environmental and visual conditions encountered in construction practice.

Existing construction image datasets, including ConstructionSite 10k~\cite{chen2025largepretrainedvisionlanguage}, SODA~\cite{duan2022soda}, ACID~\cite{xiao2021development}, MOCS~\cite{xuehui2021dataset}, and related collections, have supported research in object detection, instance segmentation, image captioning, visual question answering (VQA), scene understanding, and safety analysis. However, most existing datasets provide limited systematic representation of adverse weather, fog, nighttime conditions, and small-object or long-distance views~\cite{xiong2026openconstruction,ding2024robust}. Such conditions may be present in existing collections, but are often sparsely represented or not explicitly organized by condition type, thereby limiting condition-specific evaluation of model robustness. Systematic acquisition of comparable real-world construction imagery across predefined adverse conditions is also difficult because weather events are intermittent, nighttime construction activities may occur less frequently, site access can be constrained, and detailed annotation of construction objects and activities is resource intensive~\cite{lee2022synthetic,barrera2023generating}.

\begin{table}[htbp]
\centering
\small
\caption{Comparison of \textit{ConSynth-X} with representative construction image datasets. ``Paired'' indicates direct correspondence between synthetic images and their source images.}
\label{tab:dataset-comparison}

\begin{adjustbox}{max width=\linewidth}
\begin{tabular}{
p{2.2cm}
p{4.0cm}
p{4.0cm}
p{1.2cm}
p{1.5cm}
p{1.4cm}
}
\toprule
\textbf{Dataset} &
\textbf{Tasks} &
\textbf{Adverse conditions} &
\textbf{Paired} &
\textbf{Type} &
\textbf{Images} \\
\midrule

\textbf{ConSynth-X (ours)} &
Object detection, VQA, image captioning, Visual grounding &
11 subsets across five condition groups &
\cmark &
Synthetic &
34,199 \\

ExtCon~\cite{ding2024robust} &
Object detection &
Five adverse conditions &
\xmark &
Synthetic &
506 \\

SODA~\cite{duan2022soda} &
Object detection &
Not explicitly categorized &
\xmark &
Real &
19,846\\

ACID~\cite{xiao2021development,xiao2022captioning} &
Object detection, instance segmentation, image captioning &
Not explicitly categorized &
\xmark &
Real &
10,000 \\

MOCS~\cite{xuehui2021dataset} &
Object detection, instance segmentation &
Not explicitly categorized &
\xmark &
Real &
41,668 \\

ConstructionSite 10k~\cite{chen2025largepretrainedvisionlanguage} &
Visual grounding, VQA, image captioning &
Illumination and scale/viewpoint &
\xmark &
Real &
10,013 \\

\bottomrule
\end{tabular}
\end{adjustbox}

\end{table}

Synthetic image generation provides a scalable strategy for augmenting construction datasets with environmental and visual conditions that are difficult to capture systematically in field settings~\cite{xiong2021machine}. However, its effectiveness depends on reproducing the intended environmental effects while preserving scene semantics, object geometry, spatial relationships, and annotation integrity. This requirement is particularly important for construction imagery, which frequently contains partial occlusions, temporary structures, unfinished components, heterogeneous object classes, and substantial variation in object scale~\cite{lee2022synthetic}. Rain, snow, and fog can alter visibility, contrast, surface appearance, and occlusion through processes such as atmospheric scattering and depth-dependent attenuation~\cite{kang2022application}, while nighttime conditions introduce substantial variation in illumination, shadows, reflections, color appearance, and object visibility. Inaccurate synthetic transformations may distort salient scene content, compromise annotation validity, or introduce distributional artifacts unrelated to the intended condition. Controlled generation and systematic validation are therefore necessary to assess visual fidelity, semantic preservation, and source--synthetic consistency~\cite{jin2023development}.

\begin{figure}[ht]
\centering
\includegraphics[width=0.98\linewidth]{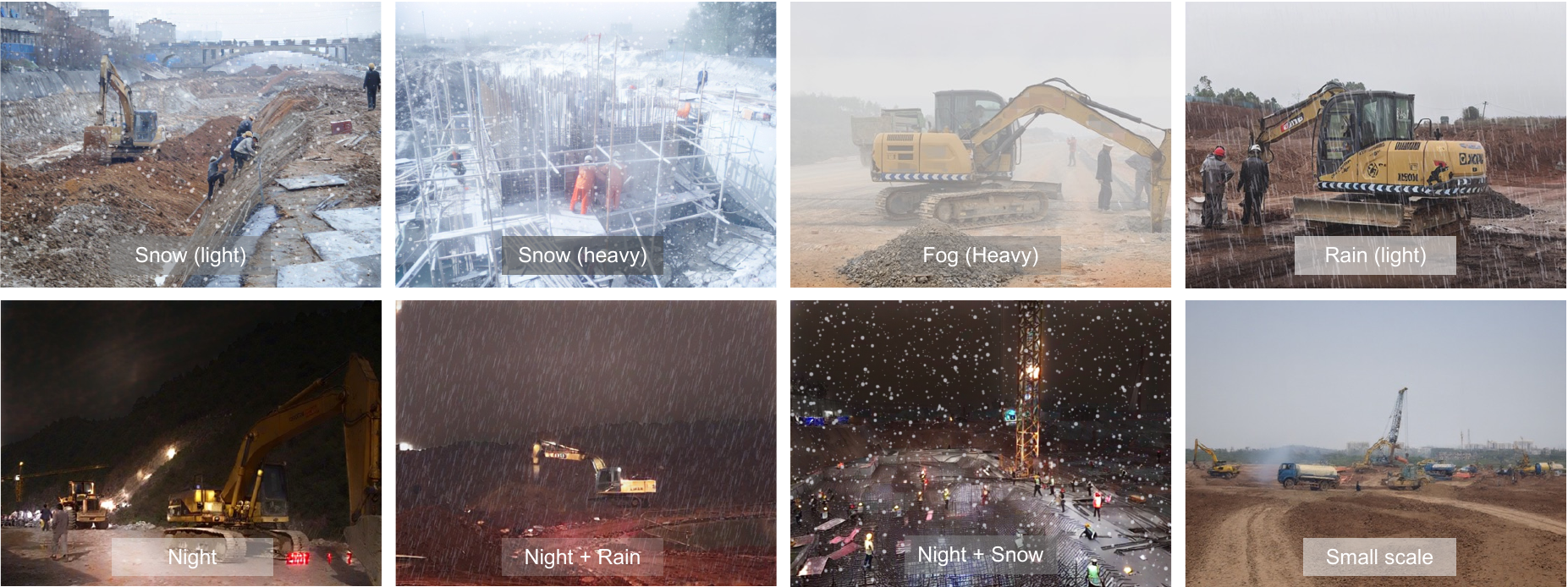}
\caption{Representative synthetic images from the \textit{ConSynth-X} dataset.}
\label{fig:condition-examples}
\end{figure}

To address these limitations, we developed \textit{ConSynth-X}, a large-scale paired synthetic construction-site image dataset for computer-vision and vision-language applications. The dataset comprises 34,199 images organized into 11 condition-specific subsets derived from real-world construction imagery. These subsets span five condition groups: precipitation, fog, nighttime conditions, adverse weather at night, and small-object or long-distance views. Each synthetic image is directly linked to its corresponding source image, enabling controlled comparison of model performance across environmental and visual conditions. Representative examples are shown in Fig.~\ref{fig:condition-examples}, and Table~\ref{tab:dataset-comparison} compares \textit{ConSynth-X} with representative construction image datasets.

Each \textit{ConSynth-X} record includes the generated image, source-image identifier, source dataset, condition label, generation method, generation parameters, source-derived annotations, and image-quality indicators. Available annotations vary by source dataset and include object bounding boxes, image captions, visual grounding annotations, and safety question--answer pairs. The dataset supports object detection, image captioning, visual grounding, and visual question answering, while the paired source--synthetic structure enables condition-specific assessment of model robustness. Technical validation includes image-quality assessment, distributional comparison with real construction imagery under corresponding adverse conditions, and source--synthetic similarity analysis to assess preservation of scene content.

\section{Methods}
\begin{figure}[htbp]
    \centering
    \includegraphics[width=0.95\linewidth]{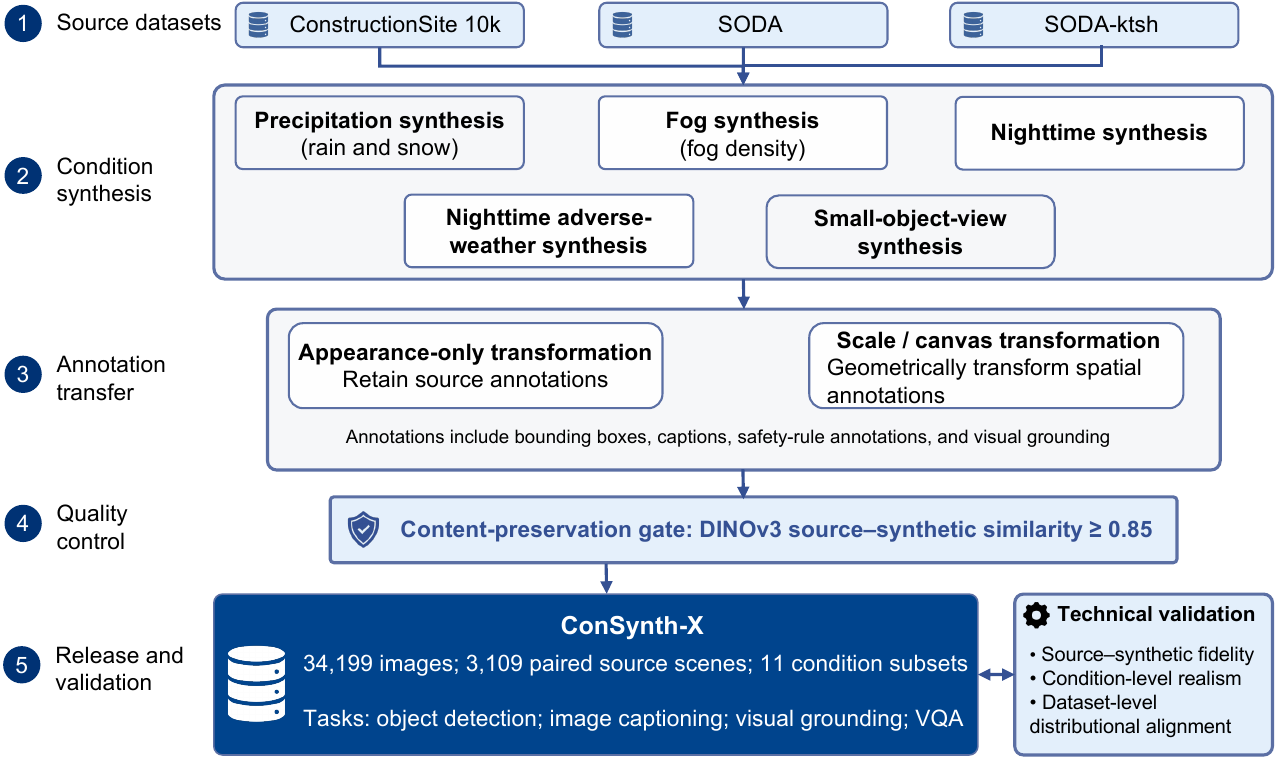}
    \caption{Overview of the \textit{ConSynth-X} generation and validation workflow.}
    \label{fig:workflow}
\end{figure}

\textit{ConSynth-X} was developed for condition-specific robustness evaluation of visual recognition models in construction-site monitoring. Real construction-site images were transformed into paired synthetic variants representing adverse weather, reduced visibility, nighttime conditions, compound nighttime-weather conditions, and small-object observation scenarios. The overall dataset generation, annotation transfer, quality control, and technical validation framework is shown in Fig.~\ref{fig:workflow}.

Five synthesis pipelines were implemented: precipitation, fog, day-to-night, nighttime adverse-weather, and small-object-view synthesis. Each generated sample maintains correspondence with its source image and associated annotations, together with condition labels, generation parameters, and quality-control metadata.

\subsection{Source data}
\textbf{Selection criteria.} Source datasets were selected based on six criteria: (1) sufficient scale for condition-specific synthesis and evaluation; (2) availability of structured annotations transferable to synthetic images; (3) coverage of construction-relevant objects and scene semantics; (4) representation of real construction environments; (5) public accessibility with terms permitting research use and attribution; and (6) predominance of clear or nominal visual conditions suitable for use as reference images in paired adverse-condition synthesis.

\textbf{Selected datasets.} Three public construction-image datasets were selected to provide complementary annotation types and task coverage. \textit{ConstructionSite 10k}~\cite{chen2025largepretrainedvisionlanguage} contains 10,013 construction-site images with bounding-box annotations for excavators, rebar, and workers with white hard hats, together with image captions and safety-rule annotations. \textit{SODA}~\cite{duan2022soda} contains 19,846 images with object-detection annotations for 15 construction-related classes covering workers, materials, equipment, and temporary site elements. For this study, 9,988 images from SODA-KTSH~\cite{deng2025enabling} were available for synthesis, spanning 16 construction-scene categories and accompanied by five reference captions per image. The original annotation structures and label vocabularies of the three source datasets were retained and linked to the corresponding \textit{ConSynth-X} records through source identifiers and dataset-specific metadata.

\subsection{Condition Schema}
The \textit{ConSynth-X} condition schema defines five environmental and observational condition groups relevant to construction-site visual perception: precipitation, fog, nighttime illumination, adverse weather at night, and small-object or long-distance views. These conditions represent variations in weather, visibility, illumination, and apparent object scale that can affect visual recognition performance~\cite{ding2024robust}. Table~\ref{tab:taxonomy-schema} summarizes the five condition groups, their primary visual effects, and the 11 condition-specific subsets included in \textit{ConSynth-X}.

\begin{table}[htbp]
\small
\centering
\caption{Adverse-condition schema used in \textit{ConSynth-X}, comprising 11 conditions across five condition groups.}
\label{tab:taxonomy-schema}

\begin{adjustbox}{max width=\linewidth}
\begin{tabular}{
p{2.9cm}
p{6.5cm}
p{3.6cm}
p{1.7cm}
}
\toprule
\textbf{Condition group} &
\textbf{Primary visual effects} &
\textbf{Conditions} &
\textbf{No. of conditions} \\
\midrule

Precipitation &
Rain or snow particles, atmospheric veiling, and changes in appearance and visibility. &
Light rain, heavy rain, light snow, heavy snow &
4 \\

Fog &
Depth-dependent reduction in visibility and contrast due to atmospheric scattering. &
Light fog, moderate fog, heavy fog &
3 \\

Low illumination &
Reduced illumination and changes in contrast, appearance, shadows, and visibility. &
Nighttime &
1 \\

Adverse weather at night &
Combined effects of low illumination and precipitation on visibility and appearance. &
Rain at night, snow at night &
2 \\

Small-object/long-distance view &
Reduced apparent object size and detail under distant or wide-area observation. &
Small-object view &
1 \\

\bottomrule
\end{tabular}
\end{adjustbox}
\end{table}

\subsection{Generation Pipeline}
Five synthesis pipelines were implemented for precipitation, fog, low illumination, adverse weather at night, and small-object/long-distance views. Each pipeline transforms a source construction image while retaining its linkage to the source image and associated annotations. The principal generation parameters and processing settings are summarized in Table~\ref{tab:generation-parameters}.

\begin{table}[htbp]
\small
\centering
\caption{Generation parameters and processing settings used in the condition-specific synthesis of \textit{ConSynth-X}.}
\label{tab:generation-parameters}
\begin{adjustbox}{max width=\linewidth}
\begin{tabular}{
P{3cm}
P{5.1cm}
P{6.5cm}
}
\toprule
\textbf{Condition group} &
\textbf{Parameter or setting} &
\textbf{Value or configuration} \\
\midrule

\multirow{12}{=}{\raggedright Precipitation}
& Source-image conditioning scale & 1.5 \\

& Text-prompt guidance scale
& 10.0 (\texttt{rain light}); 8.0 (\texttt{snow light});
12.0 (\texttt{snow heavy}) \\

& Diffusion inference steps & 30 \\

& Heavy-condition processing
& \texttt{rain heavy}: no second diffusion pass;
snow variants: shared particle overlay \\

& Rain-streak angle ($\theta$) & $75^{\circ}$--$87^{\circ}$ \\

& Rain-streak density ($n_r$; far/mid/near)
& Light: 1500--2500 / 1000--2000 / 300--700;
heavy: 3000--4500 / 2200--3600 / 900--1600 \\

& Rain-streak length ($\ell$; far/mid/near)
& Light: 10--22 / 18--35 / 25--50\,px;
heavy: 12--26 / 22--42 / 30--60\,px \\

& Rain-streak opacity ($\alpha_r$; far/mid/near)
& Light: 0.25 / 0.35 / 0.40;
heavy: 0.35 / 0.48 / 0.55 \\

& Atmospheric-veil opacity ($\alpha_a$)
& Light: 0.15--0.30; heavy: 0.30--0.45 \\

& Snowflake density ($n_s$; far/mid/near)
& 1500--3000 / 500--1000 / 100--300 \\

& Snowflake radius ($r_s$; far/mid/near)
& 1--2 / 2--4 / 4--6\,px \\

& Snowflake opacity ($\alpha_s$; far/mid/near)
& 0.35 / 0.50 / 0.65 \\

\midrule
\multirow{2}{=}{\raggedright Fog}
& Visibility distance ($V$) &
750--1000\,m (light); 500--750\,m (moderate);
300--500\,m (heavy) \\

& Depth estimation &
Depth Anything V2 (Small model variant) \\

\midrule

\multirow{2}{=}{\raggedright Low illumination}
& Translation input resolution & $512\times512$ pixels \\
& Output resampling &
Lanczos interpolation \\

\midrule

Adverse weather at night &
Transformation sequence &
Weather editing $\rightarrow$ day-to-night translation $\rightarrow$ precipitation overlay  \\

\midrule

\multirow{2}{=}{\raggedright Small-object/long-distance view}
& Canvas expansion factor ($s$) &
Truncated normal distribution: $\mu_s=0.25$, $\sigma_s=0.05$, and $0.20 \leq s \leq 0.40$ \\
& Canvas scale factor &
$1+2s$ \\

\bottomrule
\end{tabular}
\end{adjustbox}
\end{table}

\subsubsection{Precipitation synthesis}
\label{sec:weather}
Rain and snow conditions were generated using a two-stage procedure that separates global weather modification from visible precipitation rendering. In the first stage, InstructPix2Pix~\cite{brooks2023instructpix2pix} was used for text-guided editing of the source images. The rain prompt was \textit{``a rainy day with dark overcast sky, rain falling, grey clouds''}, and the snow prompt was \textit{``a cold winter day with snow, frost on surfaces, grey sky, snow on the ground''}. Source-image conditioning was used to preserve the underlying scene content while modifying weather-related appearance. The generation parameters are summarized in Table~\ref{tab:generation-parameters}.

In the second stage, visible precipitation was rendered using a three-layer physics-based particle overlay representing far-, mid-, and near-field rain streaks or snowflakes. The heavy rain variant was produced by applying an additional denser precipitation and atmospheric veil overlay to the accepted light rain image, without a second diffusion pass. The light and heavy snow variants used different diffusion prompts and guidance scales but shared the same three-layer particle overlay. The corresponding generation parameters are reported in Table~\ref{tab:generation-parameters}.

\subsubsection{Fog synthesis}
Fog conditions were generated using the Koschmieder atmospheric scattering model~\cite{narasimhan2002vision}:
\begin{equation} 
\label{eq:koschmieder} 
I_{\mathrm{fog}}(\mathbf{x}) = I(\mathbf{x})e^{-\beta d(\mathbf{x})} + A\left(1-e^{-\beta d(\mathbf{x})}\right), 
\end{equation}
where $I(\mathbf{x})$ is the source-image intensity at location $\mathbf{x}$, $d(\mathbf{x})$ is the estimated scene depth, $A$ denotes atmospheric light, and $\beta$ is the atmospheric scattering coefficient. 

Scene depth was estimated using Depth Anything V2~\cite{yang2024depth} with the Small model variant. Fog intensity was controlled through condition-specific visibility ranges of 750--1000 m (light), 500--750 m (moderate), and 300--500 m (heavy), from which the atmospheric attenuation was determined in the Koschmieder model. The transformation modifies image visibility and contrast as a function of estimated scene depth without altering the image coordinate system.

\subsubsection{Low-illumination synthesis}
Low-illumination conditions were generated through CycleGAN-Turbo-based day-to-night translation~\cite{parmar2024one}. Source images were resized to the model input resolution, translated to nighttime appearance, and resampled to their original dimensions using Lanczos interpolation. The translation modifies illumination, color balance, contrast, and shadow characteristics without explicitly changing the image coordinate system.

\subsubsection{Adverse weather at night}
Rain-at-night and snow-at-night conditions were generated using a three-stage sequence. First, InstructPix2Pix was applied to the daytime source image to synthesize the corresponding weather appearance. Second, the weather-modified image was translated to nighttime appearance using CycleGAN-Turbo. Finally, the corresponding rain or snow particle overlay was applied to render visible precipitation.

\subsubsection{Small-object/long-distance-view synthesis}
Small-object/long-distance-view conditions were generated through canvas expansion and FLUX.1-based image outpainting~\cite{blackforestlabs2024fluxfill}. Each source image was placed within an expanded canvas, and the newly exposed regions were synthesized using the prompt \textit{``Extend the image edges seamlessly. Continue only the existing ground texture, dirt, concrete, and sky. Match lighting, colors, and perspective. Do not add any new objects''}. The procedure reduces the relative image area occupied by the original scene, representing distant or wide-area observations.

The per-side canvas-expansion factor $s$ was sampled from a normal distribution constrained to the interval $[0.20,0.40]$:

\begin{equation} 
\label{eq:scale-factor} 
s \sim \mathcal{N}(\mu_s=0.25,\sigma_s=0.05), \qquad s\in[0.20,0.40]. \end{equation}
where the expanded canvas size is $(1 + 2s) \times$ the original dimension along each axis.

\subsection{Dataset Annotation}
\label{sec:annotation}
\textit{ConSynth-X} retains the annotations associated with each source image, including object bounding boxes, image captions, and safety-related annotations, depending on the source dataset. Each generated image is linked to its corresponding source image through a unique source identifier, enabling source-derived annotations to be propagated to the synthetic variants. The detailed annotation fields and storage schema are described in the Data Records section.

\begin{figure}[ht]
    \centering
    \includegraphics[width=0.95\linewidth]{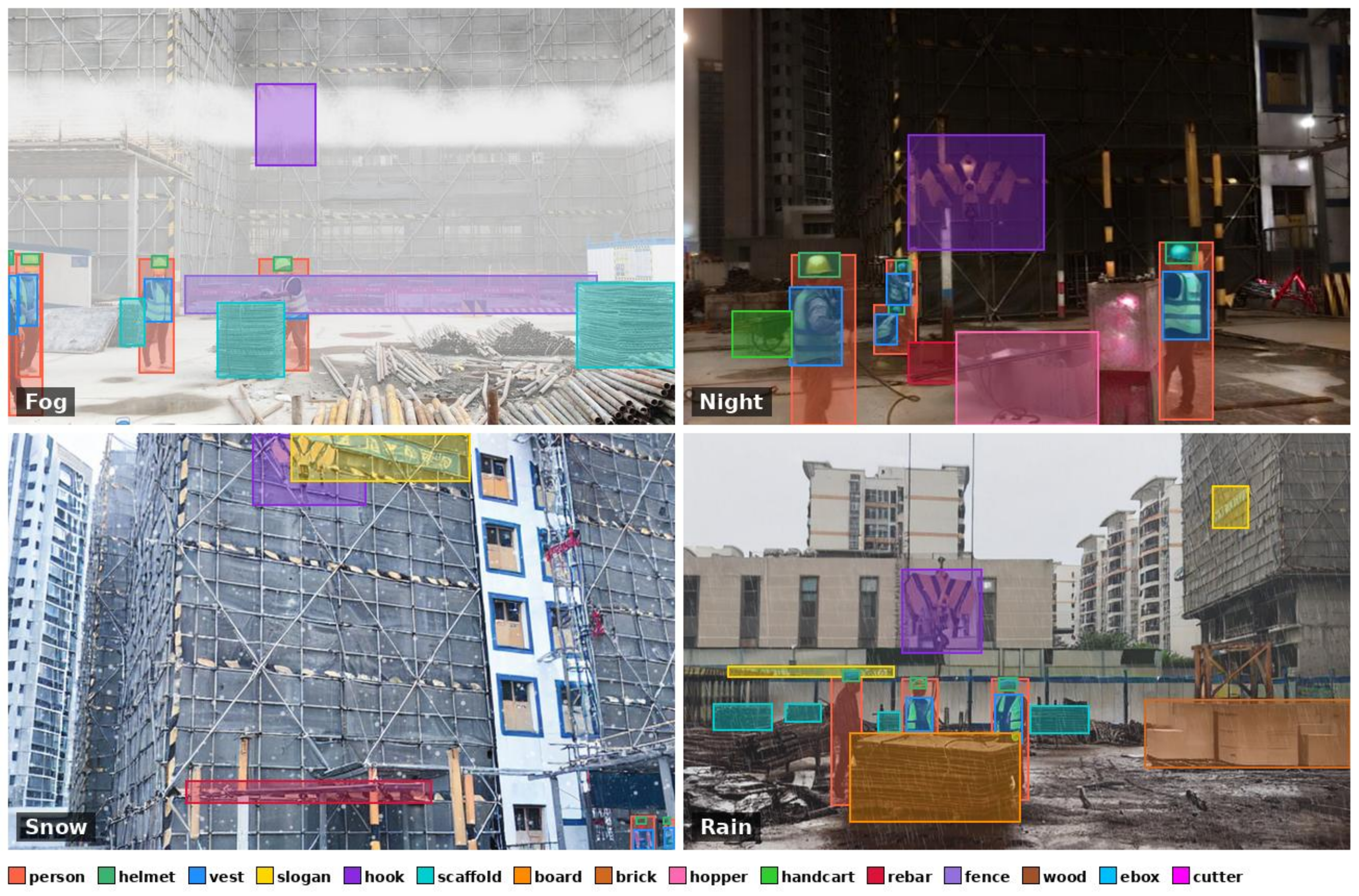}
    \caption{Representative bounding-box annotations transferred across fog, nighttime, snow, and rain conditions in \textit{ConSynth-X}.}
    \label{fig:annotations}
\end{figure}

Annotation transfer depends on whether the synthesis procedure changes image geometry. For precipitation, fog, low-illumination, and adverse-weather-at-night conditions, the image coordinate system is unchanged; therefore, spatial annotations are transferred directly from the corresponding source images. Non-spatial annotations, including captions and textual safety annotations, are retained without modification. Fig.~\ref{fig:annotations} shows representative bounding-box annotations across appearance-modified conditions.

For the small-object/long-distance-view condition, the source image is resized and positioned within an expanded canvas, requiring spatial annotations to be transformed to the new coordinate system. Let $(W,H)$ denote the source-image dimensions, $r$ the image-scaling factor, $(p_x,p_y)$ the placement offset, and $(W',H')$ the expanded-canvas dimensions. For a normalized source bounding box $(x_1,y_1,x_2,y_2)$, the transformed coordinates are

\begin{equation}
\label{eq:bbox-transfer}
\begin{aligned}
x'_1 &= \frac{x_1Wr+p_x}{W'}, &
y'_1 &= \frac{y_1Hr+p_y}{H'},\\[2pt]
x'_2 &= \frac{x_2Wr+p_x}{W'}, &
y'_2 &= \frac{y_2Hr+p_y}{H'}.
\end{aligned}
\end{equation}
where $(x'_1,y'_1,x'_2,y'_2)$ are the normalized bounding-box coordinates in the expanded canvas.

\section{Data Records}
The \textit{ConSynth-X} dataset is deposited on Hugging Face~\cite{duong2026consynthx}. The release contains 34,199 synthetic image records derived from 3,109 retained source scenes across three construction image datasets: ConstructionSite 10k, SODA, and SODA-KTSH. Each retained source scene is represented under each of the 11 adverse-condition subsets, resulting in 17,446 records derived from 1,586 ConstructionSite 10k source scenes, 6,688 records derived from 608 SODA source scenes, and 10,065 records derived from 915 SODA-KTSH source scenes.

The 11 subsets correspond to light rain, heavy rain, light snow, heavy snow, light fog, moderate fog, heavy fog, nighttime, rain at night, snow at night, and small-object/long-distance view. Records are organized by source dataset and adverse-condition label, enabling condition-specific analysis as well as paired comparison of the same source scene across multiple environmental and visual conditions. The source--synthetic linkage is preserved through a source-image identifier associated with every synthetic record.

The three source datasets provide complementary annotation types. ConstructionSite 10k contributes object bounding boxes, image captions, and safety-related annotations; SODA contributes object-detection annotations for 15 construction-related classes; and SODA-KTSH contributes five reference captions per image. These source-derived annotations are retained or geometrically transformed as described in Section~\ref{sec:annotation}, depending on whether the corresponding synthesis operation changes the image coordinate system.

\begin{figure}[htbp]
    \centering
    \includegraphics[width=1\linewidth]{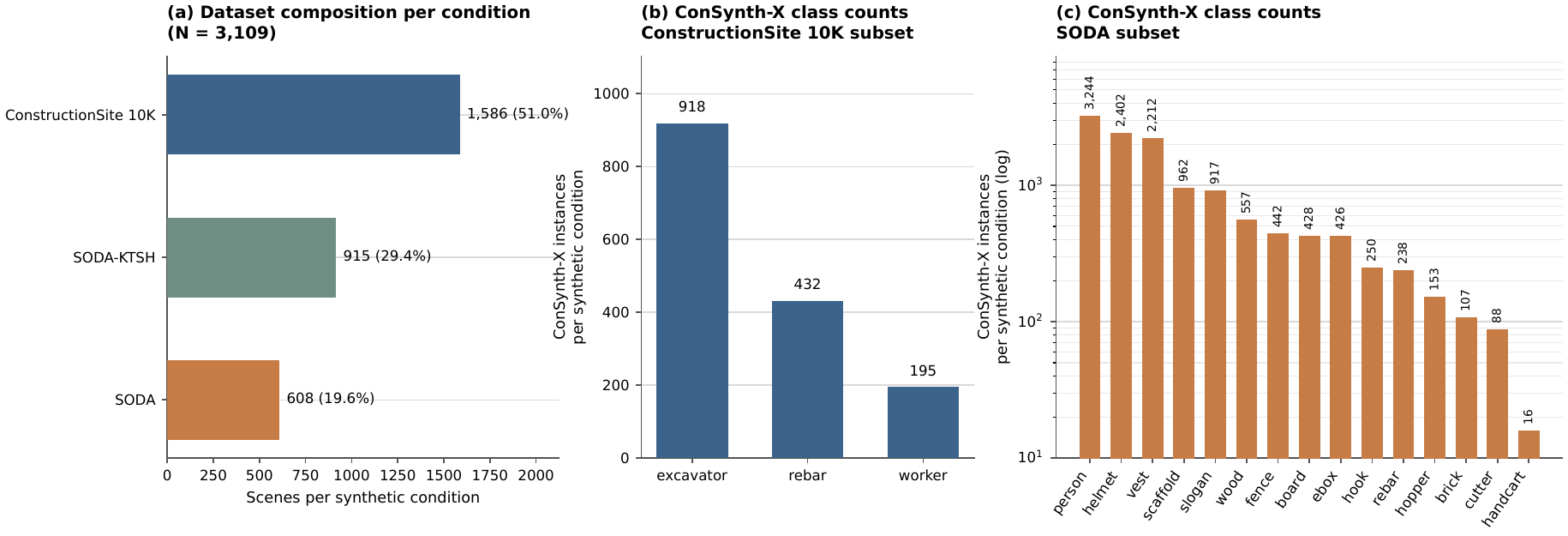}
    \caption{Dataset composition and inherited object annotations in \textit{ConSynth-X}.}
    \label{fig:data_distribute}
\end{figure}

Fig.~\ref{fig:data_distribute} summarizes the contribution of each source dataset and the class composition of the inherited object-detection annotations. Because the same retained source scenes are represented across all 11 synthetic conditions, the source-dataset proportions and inherited class distributions remain constant across condition subsets. SODA-KTSH is caption-based and therefore does not contribute object-detection classes to Fig.~\ref{fig:data_distribute}.

All synthetic records are distributed in Parquet format. Each record contains the generated image, source-image identifier, source-dataset identifier, adverse-condition label, source-derived annotations, generation metadata, and quality-assessment results. Generated images are stored as JPEG-encoded binary data. Where object-detection annotations are available, bounding boxes are represented using normalized $[x_1,y_1,x_2,y_2]$ coordinates in the interval $[0,1]$. The principal data elements are summarized in Table~\ref{tab:data-schema}.

\begin{table}[ht]
\centering
\small
\caption{Principal data elements provided for each \textit{ConSynth-X} record.}
\label{tab:data-schema}

\begin{adjustbox}{max width=\linewidth}
\begin{tabular}{
P{3.3cm}
P{3.0cm}
P{8.0cm}
}
\toprule
\textbf{Data element} &
\textbf{Format} &
\textbf{Description} \\
\midrule

Generated image &
JPEG-encoded binary &
Synthetic construction-site image corresponding to the specified adverse-condition subset. \\

Source-image identifier &
String identifier &
Unique identifier linking each synthetic record to its corresponding source scene. \\

Source dataset &
Categorical &
Dataset of origin for the source scene. \\

Condition label &
Categorical &
Adverse-condition category assigned to the synthetic record. \\

Source-derived annotations &
Dataset-specific structured fields &
Annotations propagated from the corresponding source dataset. \\

Object bounding boxes &
Normalized coordinates &
Object locations represented as $[x_1,y_1,x_2,y_2]$ coordinates normalized to $[0,1]$.\\
Generation metadata &
Condition-specific metadata &
Parameters and processing information associated with the synthesis procedure. \\

Quality-assessment metrics &
Numeric &
Per-image measures of source--synthetic correspondence and image quality. \\

\bottomrule
\end{tabular}
\end{adjustbox}
\end{table}

The paired organization of the dataset allows users to retrieve synthetic variants of the same source scene across conditions and to compare model performance between corresponding source and synthetic images. The detailed release structure, field definitions, and access information are provided with the deposited dataset and are summarized in the Data Availability statement.

\section{Data Overview}
\begin{figure}[htbp]
    \centering
    \includegraphics[width=1\linewidth]{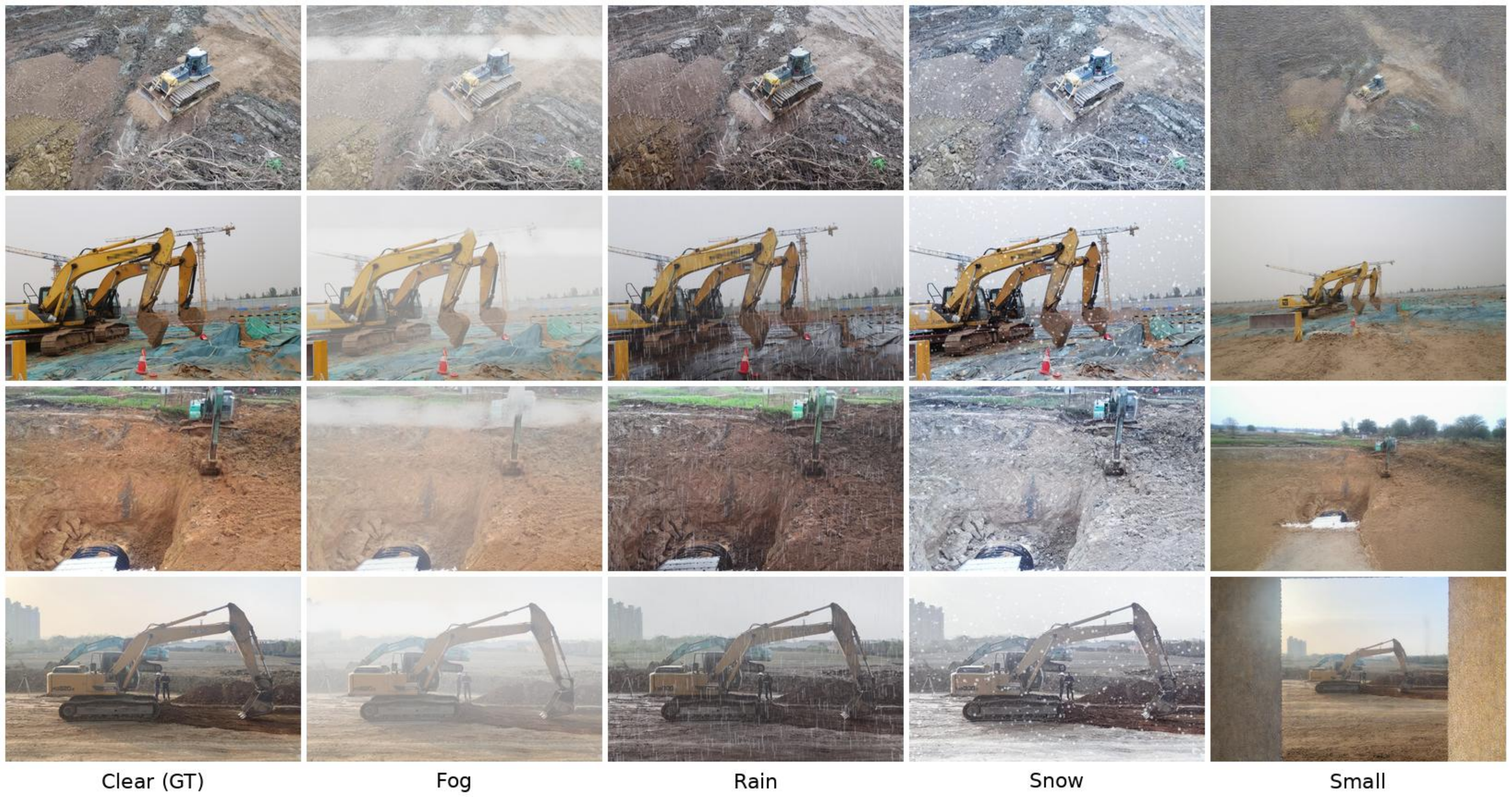}
    \caption{Representative source images and corresponding synthetic variants in \textit{ConSynth-X}.}
    \label{fig:paired-examples}
\end{figure}

Fig.~\ref{fig:paired-examples} presents representative source--synthetic image pairs from \textit{ConSynth-X}. Each row corresponds to a single construction scene, with the source image and its synthetic variants shown across different environmental and visual conditions. The paired organization preserves scene correspondence across conditions and facilitates direct comparison of condition-specific visual changes.

\section{Technical Validation}
Technical validation comprised three complementary analyses: (1) paired source–synthetic fidelity, assessing preservation of scene content and spatial structure; (2) condition-level reference alignment, assessing whether synthetic images move closer than their source images to real adverse-condition reference distributions; and (3) dataset-level distributional alignment, comparing synthetic and source-image feature distributions with real adverse-condition reference sets using FID and KID. Together, these analyses evaluate source-content preservation and feature-space alignment of the synthesized conditions.

\subsection{Source--Synthetic Fidelity}
Source--synthetic fidelity was evaluated using DINOv3~\cite{simeoni2025dinov3}, a self-supervised vision foundation model that produces semantically informative image representations without task-specific supervision. DINO-family representations capture high-level semantic and spatial information while remaining relatively robust to low-level appearance variation~\cite{caron2021emerging}. This property is appropriate for evaluating environmental image synthesis, where weather and illumination characteristics are intentionally modified while scene content, object identity, and spatial configuration should remain preserved. For source image $I_i^{\mathrm{src}}$ and its synthetic counterpart $I_{i,c}^{\mathrm{syn}}$ under condition $c$, fidelity was quantified by the cosine similarity

\begin{equation}
\label{eq:dino-cosine}
s_{i,c}
=
\frac{
f(I_i^{\mathrm{src}})^{\top}
f(I_{i,c}^{\mathrm{syn}})
}{
\left\|f(I_i^{\mathrm{src}})\right\|_2
\left\|f(I_{i,c}^{\mathrm{syn}})\right\|_2
}.
\end{equation}

where $f(\cdot)$ denotes the DINOv3 image encoder. DINO-based similarity has also been used as a measure of subject fidelity in generative image synthesis~\cite{ruiz2023dreambooth}. SSIM, LPIPS, and CLIP similarity scores are additionally provided in the dataset metadata as complementary measures of source--synthetic correspondence.

A minimum DINOv3 cosine similarity of 0.85 was applied to eight synthetic subsets representing light rain, light and heavy snow, three fog intensity levels, nighttime, and small-object conditions. The criterion was applied jointly at the source-scene level: a source scene was retained only when its synthetic images satisfied the threshold for all eight subsets. The threshold was selected based on the empirical similarity distributions and the corresponding sample-retention analysis. The sensitivity of sample retention to the threshold is reported in Table~\ref{tab:dino-sweep} in the Appendix.

At the selected threshold, 3,109 source scenes satisfied the joint fidelity criterion. Consequently, every synthetic image in the threshold-controlled subsets has a DINOv3 cosine similarity of at least 0.85 with its corresponding source image. The criterion provides a consistent semantic and fidelity requirement while permitting the intended changes in weather, illumination, and object scale.

The fidelity criterion was not applied to the subsets representing heavy rain, rain at night, and snow at night. These three subsets were retained as part of the predefined condition set but were evaluated independently of the admission threshold. Their per-image DINOv3 similarity scores are included in the dataset metadata to support condition-specific analysis and filtering.

\subsection{Condition-Level Reference Alignment}
Condition-level reference alignment was evaluated by comparing the distances of each synthetic image and its corresponding source image to a real-image reference distribution representing the same adverse condition in DINOv3 embedding space. 

For each source--synthetic pair, the change in distance was defined as

\begin{equation} 
\Delta d = d(\mathbf{z}_{\mathrm{src}}, \mathcal{R}_{c}) - d(\mathbf{z}_{\mathrm{syn}}, \mathcal{R}_{c}), 
\end{equation}
where $\mathbf{z}_{\mathrm{src}}$ and $\mathbf{z}_{\mathrm{syn}}$ denote the DINOv3 embeddings of the source and synthetic images, respectively, and $\mathcal{R}_{c}$ denotes the reference set for condition $c$. The distance $d(\mathbf{z},\mathcal{R}_{c})$ was computed as the Mahalanobis distance between embedding $\mathbf{z}$ and the reference distribution characterized by its mean and shrinkage-estimated covariance matrix. Covariance shrinkage was set to 0.10.

Condition-specific changes were summarized using paired Cohen's $d_z$ and the proportion of source--synthetic pairs with $\Delta d>0$.  A positive $\Delta d$ indicates that the synthetic image is closer than its corresponding source image to the real-condition reference distribution.

The analysis was conducted using the ConstructionSite 10k-derived subset, which contains 1,586 retained source scenes and represents the largest source subset in \textit{ConSynth-X}. Real-image references for fog and snow were drawn from ACDC, WeatherBench, and WeatherNet~\cite{sakaridis2021acdc,guan2025weatherbench,weathernet_hf}. Nighttime references were obtained from ACDC, while rain references were drawn from ACDC and WeatherBench. An additional 199 construction-site rain images were retrieved through the Openverse API~\cite{openverse2026} and used as construction-domain references for the synthetic rain conditions. These images were screened for both construction-site content and visible precipitation using a two-stage CLIP-based zero-shot classification procedure. The small-object condition was excluded from the reference-based analysis due to the absence of an appropriate real-image reference set.

\begin{figure}[htbp] 
\centering 
\includegraphics[width=0.85\linewidth]{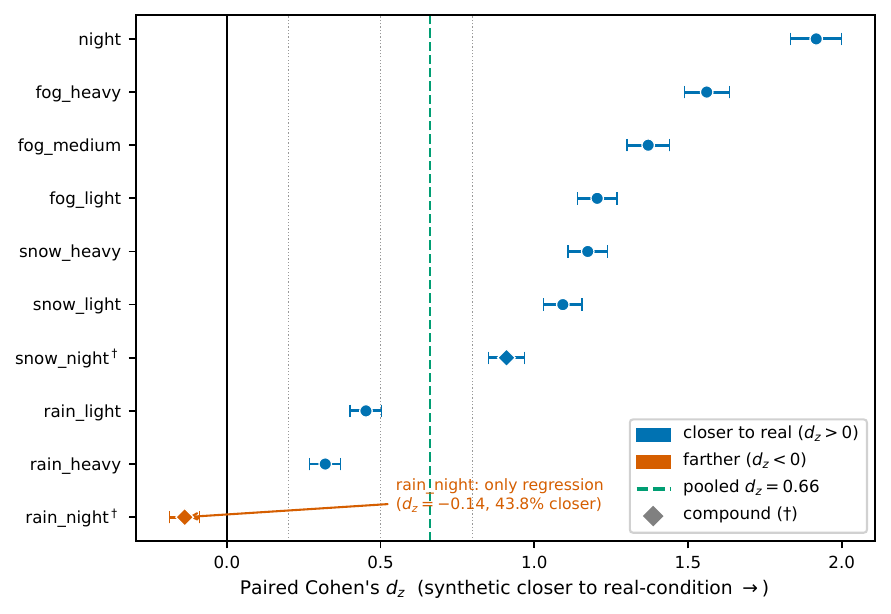} \caption{Condition-level reference alignment assessed using paired DINOv3 embedding distances to real adverse-condition reference distributions.} 
\label{fig:dino-drift} 
\end{figure}

Positive distance differences were observed in nine of the ten evaluated conditions (Fig.~\ref{fig:dino-drift}), with the largest standardized shifts for nighttime, fog, and snow. Overall, 81.85\% of synthetic images were closer to their corresponding real-condition reference distributions than the source images, yielding a pooled paired effect size of $d_z=0.6615$. The rain-at-night condition was the only condition with a negative effect size ($d_z=-0.1374$), with synthetic images showing slightly greater distances to the corresponding real-condition reference distribution than their source counterparts.

\subsection{Dataset-Level Distributional Alignment}
\label{subsec:fid}
Dataset-level distributional alignment was evaluated using Fr\'echet Inception Distance (FID)~\cite{heusel2017gans} and Kernel Inception Distance (KID)~\cite{binkowski2018demystifying}. Both metrics were computed from 2,048-dimensional InceptionV3 pool3 features. KID was estimated using subsets of 500 images. Lower FID and KID values indicate greater similarity between the feature distributions of the synthetic images and the corresponding real adverse-condition reference set.

ConstructionSite 10k, SODA, and SODA-KTSH were evaluated separately to account for differences in source-image composition and sample size. Fog, rain, and snow conditions were compared with real-image reference sets from ACDC~\cite{sakaridis2021acdc}, WeatherBench~\cite{guan2025weatherbench}, and WeatherNet~\cite{weathernet_hf}. Nighttime conditions were evaluated against ACDC. The small-object condition was excluded because no directly comparable real-image reference distribution was available. 

For each source dataset, adverse condition, and reference set, the corresponding source images were evaluated as the baseline. Distributional changes were therefore assessed by comparing each synthetic subset with its source-image baseline under the same reference set. FID values were not compared across source datasets because FID is sensitive to sample size and dataset composition. KID was included as a complementary measure with reduced finite-sample bias~\cite{binkowski2018demystifying}.

\begin{table}[htbp]
\centering
\small
\caption{Distributional alignment of the ConstructionSite 10k subset with real adverse-condition reference sets.}
\label{tab:fid-cs10k}

\begin{tabular}{lccc}
\toprule
\textbf{Condition} &
\textbf{ACDC} &
\textbf{WeatherBench} &
\textbf{WeatherNet} \\
\midrule

\textit{Source (fog)}  & 220.1 / 0.199 & 217.5 / 0.111 & 161.7 / 0.089 \\
Light fog              & 214.4 / 0.196 & 206.2 / 0.106 & 148.1 / 0.080 \\
Moderate fog           & 214.4 / 0.196 & 202.1 / 0.103 & 144.9 / 0.078 \\
Heavy fog              & 217.1 / 0.198 & 193.8 / 0.097 & 140.8 / 0.076 \\

\midrule

\textit{Source (rain)} & 199.3 / 0.176 & 244.3 / 0.116 & 159.4 / 0.088 \\
Light rain             & 192.2 / 0.173 & 237.1 / 0.115 & 153.1 / 0.087 \\
Heavy rain             & 237.6 / 0.237 & 235.8 / 0.140 & 178.2 / 0.126 \\
Rain at night          & 203.8 / 0.197 & 227.2 / 0.140 & 164.9 / 0.124 \\

\midrule

\textit{Source (snow)} & 217.4 / 0.200 & 225.4 / 0.101 & 153.3 / 0.091 \\
Light snow             & 199.6 / 0.181 & 210.7 / 0.088 & 130.4 / 0.067 \\
Heavy snow             & 192.0 / 0.175 & 210.7 / 0.090 & 121.1 / 0.060 \\
Snow at night          & 226.9 / 0.217 & 226.4 / 0.127 & 154.0 / 0.109 \\

\midrule

\textit{Source (night)} & 282.2 / 0.304 & -- & -- \\
Nighttime               & 191.4 / 0.191 & -- & -- \\

\bottomrule
\end{tabular}
\end{table}

\begin{figure}[htbp] 
\centering 
\includegraphics[width=0.9\linewidth]{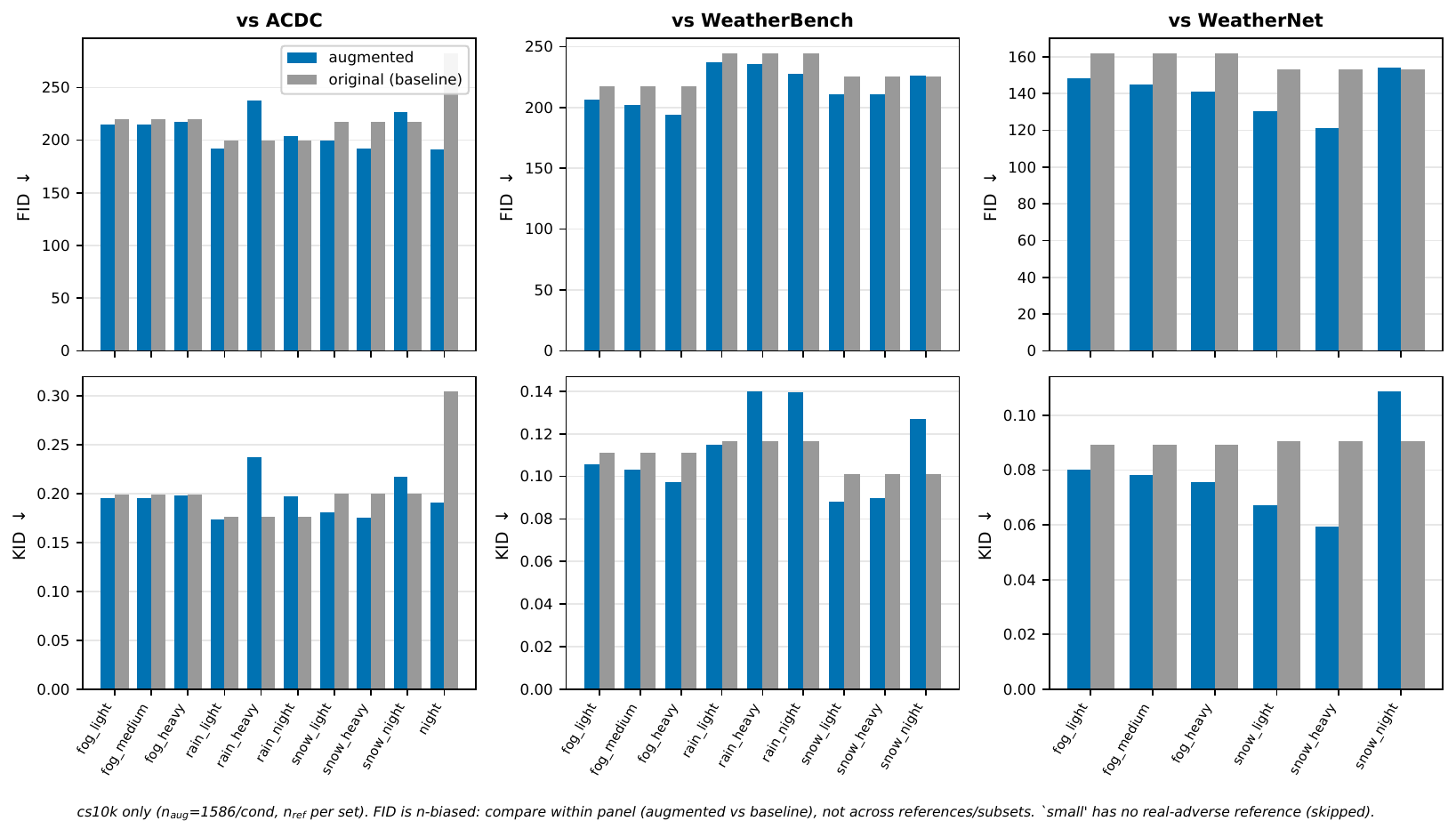} \caption{FID and KID values for the ConstructionSite 10k synthetic subsets relative to real adverse-condition reference sets.} 
\label{fig:fid-kid} 
\end{figure}

Most synthetic conditions in ConstructionSite 10k exhibited lower FID and KID than their corresponding source-image baselines (Table~\ref{tab:fid-cs10k}; Fig.~\ref{fig:fid-kid}). Improvements were consistently observed across the three fog intensity levels and for nighttime conditions. Light rain and the two daytime snow conditions also generally showed improved alignment across the available reference sets. For nighttime images, FID relative to ACDC decreased from 282.2 for the source-image baseline to 191.4 for the synthetic subset, with a corresponding decrease in KID from 0.304 to 0.191.

Heavy rain and precipitation under nighttime conditions exhibited less consistent results across reference sets. Heavy rain showed increased FID and KID relative to the source baseline for ACDC and WeatherNet, whereas its FID relative to WeatherBench decreased slightly. Rain at night and snow at night similarly showed reference-dependent behavior. These results are consistent with the weaker condition-level alignment observed for heavy rain and the negative DINOv3 effect size observed for rain at night. The SODA and SODA-KTSH subsets exhibited comparable overall trends; their complete FID and KID results are reported in Table~\ref{tab:fid-soda} in the Appendix.

\section{Usage Notes}
\textit{ConSynth-X} is released under the Creative Commons Attribution--NonCommercial 4.0 International license (CC BY-NC 4.0). The original licensing terms of the source datasets remain applicable, and users are responsible for ensuring compliance with the requirements.

The dataset is designed for non-commercial research on the robustness of construction computer-vision and vision--language models under adverse environmental and visual conditions. The represented conditions include precipitation, fog, nighttime illumination, adverse weather at night, and small-object or long-distance views. The available annotations support evaluation tasks including object detection, visual grounding, image captioning, and visual question answering, depending on the annotations provided by the corresponding source dataset.

Benchmarking should be conducted at the condition level using the source-derived annotations associated with each image pair. Object detection can be evaluated using standard detection metrics for ConstructionSite 10k and SODA, while image captioning can be assessed against the reference captions provided by ConstructionSite 10k and SODA-KTSH. Safety-oriented visual question answering on ConstructionSite 10k can be evaluated using answer accuracy, and visual grounding can be assessed using bounding-box localization performance. Because each synthetic image is paired with its corresponding source image, the dataset additionally supports direct comparison of model performance between source and synthetic images representing the same scene.

The adverse conditions represented in \textit{ConSynth-X} are synthetically generated and therefore do not reproduce the full variability of field-captured environmental conditions. The dataset is intended to complement, rather than replace, evaluation using real adverse-condition imagery. The DINOv3 fidelity criterion was applied to eight synthetic subsets to control source--synthetic consistency, whereas heavy rain, rain at night, and snow at night were retained without threshold-based filtering. Per-image DINOv3 similarity scores are provided for all conditions to facilitate additional quality assessment and sensitivity analysis.

\section{Data Availability}
The ConSynth-X dataset is deposited on Hugging Face \cite{duong2026consynthx}. The release contains 34,199 synthetic construction-site image records derived from 3,109 source scenes, organized into 11 adverse-condition subsets spanning precipitation, fog, nighttime illumination, adverse weather at night, and small-object/long-distance views. The data are distributed in Parquet format and are organized by source dataset and condition. Each record includes the generated image, source-image identifier, source dataset, condition label, source-derived annotations, generation metadata, and quality-assessment results, enabling direct linkage between synthetic images and their corresponding source scenes.

The dataset incorporates secondary data derived from ConstructionSite 10k, SODA, and SODA-KTSH. The original licensing and access terms of these source datasets remain applicable to source-derived content, and users are responsible for complying with the corresponding upstream requirements. Generation and validation code is available separately as described in the Code Availability section.

\section{Code Availability}
The code used for dataset generation and technical validation is publicly available at \url{https://github.com/ruoxinx/ConSynth-X-Dataset} under the Apache License 2.0.

\section*{Acknowledgements}
The authors gratefully acknowledge the creators of the ConstructionSite 10k, SODA, and SODA-KTSH datasets for making these resources available. The authors also acknowledge the Ohio Supercomputer Center for providing computational resources used in this study. This work was supported in part by the Farris Family Innovation Award.

\appendix
\section{Appendix}
\label{sec:supp}
\subsection*{Sensitivity Analysis of the DINOv3 Fidelity Threshold}
The sensitivity of sample retention to the DINOv3 fidelity threshold was evaluated at candidate values ranging from 0.70 to 0.90. For each threshold $\tau$, a source scene was retained only when all eight synthetic images subject to the fidelity criterion satisfied

\begin{equation} 
\min_{c \in \mathcal{C}_{\mathrm{fidelity}}} s_{i,c} \geq \tau, \end{equation} 
where $\mathcal{C}_{\mathrm{fidelity}}$ denotes the eight synthetic subsets included in the source--synthetic fidelity criterion. The three additional subsets representing heavy rain, rain at night, and snow at night were not subject to the fidelity criterion.

Table~\ref{tab:dino-sweep} reports the number of source scenes satisfying the joint criterion at each candidate threshold. At the selected threshold of 0.85, 3,109 source scenes were retained, comprising 1,586 from ConstructionSite 10k, 608 from SODA, and 915 from SODA-KTSH. These scenes correspond to 34,199 synthetic images across the 11 condition subsets. Increasing the threshold to 0.90 reduced the retained set to 688 source scenes, whereas lower thresholds retained substantially larger numbers of scenes.

\begin{table}[htbp]
\centering
\small
\caption{Sensitivity of source-scene retention to the DINOv3 fidelity threshold.}
\label{tab:dino-sweep}

\begin{tabular}{lrrrr}
\toprule
& \multicolumn{4}{c}{\textbf{Retained source scenes, $n$}} \\
\cmidrule(lr){2-5}
\textbf{DINOv3 threshold} &
\textbf{ConstructionSite 10k} &
\textbf{SODA} &
\textbf{SODA-KTSH} &
\textbf{Total} \\
\midrule
0.70 & 3,995 & 3,214 & 2,618 & 9,827 \\
0.75 & 3,485 & 2,545 & 2,284 & 8,314 \\
0.80 & 2,738 & 1,635 & 1,694 & 6,067 \\
0.85\textsuperscript{a} & 1,586 & 608 & 915 & 3,109 \\
0.90 & 406 & 65 & 217 & 688 \\
\bottomrule
\end{tabular}

\medskip
\footnotesize
\textsuperscript{a} Selected threshold used to construct the \textit{ConSynth-X} dataset.
\end{table}

\subsection*{Additional Distributional Alignment Results}
Table~\ref{tab:fid-soda} reports the distributional alignment results for the two additional source datasets using the same FID and KID procedures applied in the primary analysis. Each synthetic condition is evaluated relative to the source-image baseline from the same source dataset, adverse condition, and real-image reference set. Comparisons across source datasets are not used because of differences in sample size and source-domain composition.

\begin{table}[ht]
\centering
\small
\caption{Distributional alignment results for the additional source datasets. Values are reported as FID / KID; lower values indicate closer distributional alignment.}
\label{tab:fid-soda}

\begin{tabular}{lccc}
\toprule
\textbf{Condition} &
\textbf{ACDC} &
\textbf{WeatherBench} &
\textbf{WeatherNet} \\
\midrule

\multicolumn{4}{l}{\textbf{SODA} ($n=608$ per synthetic condition)} \\

\textit{Source (fog)}   & 259.9 / 0.220 & 261.7 / 0.161 & 206.6 / 0.117 \\
Light fog               & 244.2 / 0.206 & 237.3 / 0.143 & 184.4 / 0.101 \\
Moderate fog            & 245.4 / 0.205 & 231.3 / 0.136 & 183.8 / 0.100 \\
Heavy fog               & 252.5 / 0.213 & 222.1 / 0.129 & 186.0 / 0.103 \\

\midrule

\textit{Source (rain)}  & 222.3 / 0.173 & 275.9 / 0.141 & -- \\
Light rain              & 208.2 / 0.160 & 269.7 / 0.140 & 195.0 / 0.105\\
Heavy rain              & 225.6 / 0.195 & 270.8 / 0.161 & 200.1 / 0.127 \\
Rain at night           & 207.3 / 0.179 & 249.7 / 0.143 & 188.5 / 0.121\\

\midrule

\textit{Source (snow)}  & 243.8 / 0.208 & 261.2 / 0.134 & 206.6 / 0.122 \\
Light snow              & 231.9 / 0.193 & 254.4 / 0.128 & 188.6 / 0.104 \\
Heavy snow              & 226.0 / 0.184 & 252.3 / 0.127 & 181.6 / 0.096 \\
Snow at night           & 258.8 / 0.239 & 262.6 / 0.160 & 212.8 / 0.151 \\

\midrule

\textit{Source (night)} & 318.0 / 0.318 & -- & -- \\
Nighttime               & 249.3 / 0.241 & -- & -- \\

\midrule

\multicolumn{4}{l}{\textbf{SODA-KTSH} ($n=915$ per synthetic condition)} \\

\textit{Source (fog)}   & 256.6 / 0.232 & 246.9 / 0.160 & 188.8 / 0.114 \\
Light fog               & 252.7 / 0.228 & 236.2 / 0.154 & 178.2 / 0.108 \\
Moderate fog            & 253.8 / 0.228 & 232.5 / 0.152 & 177.3 / 0.108 \\
Heavy fog               & 256.0 / 0.229 & 223.2 / 0.142 & 176.0 / 0.107 \\

\midrule

\textit{Source (rain)}  & 223.4 / 0.190 & 261.6 / 0.135 & -- \\
Light rain              & 214.0 / 0.187 & 252.9 / 0.136 & 162.1 / 0.097\\
Heavy rain              & 244.8 / 0.234 & 248.7 / 0.149 & 183.6 / 0.128\\
Rain at night           & 209.8 / 0.193 & 231.9 / 0.139 & 160.4 / 0.112\\

\midrule

\textit{Source (snow)}  & 255.5 / 0.235 & 244.4 / 0.128 & 182.8 / 0.118 \\
Light snow              & 234.0 / 0.213 & 228.3 / 0.111 & 148.8 / 0.084 \\
Heavy snow              & 230.8 / 0.209 & 229.6 / 0.114 & 142.3 / 0.078 \\
Snow at night           & 259.9 / 0.265 & 246.5 / 0.162 & 173.4 / 0.137 \\

\midrule

\textit{Source (night)} & 316.4 / 0.332 & -- & -- \\
Nighttime               & 242.9 / 0.238 & -- & -- \\

\bottomrule
\end{tabular}
\end{table}

Across the two additional source datasets, fog conditions consistently exhibited lower FID and KID than their corresponding source-image baselines across the available reference sets. Daytime snow conditions and nighttime images also generally showed improved distributional alignment. Heavy rain and precipitation under nighttime conditions exhibited greater variability, consistent with the ConstructionSite 10k results. These findings indicate that the principal distributional trends observed in the primary analysis are also present across the additional source datasets.

\bibliographystyle{unsrtnat}
\bibliography{references}

\end{document}